# Uncertain Nearest Neighbor Classification

Fabrizio Angiulli, Fabio Fassetti
DEIS, Università della Calabria
Via P. Bucci, 41C
87036 Rende (CS), Italy
{f.angiulli,f.fassetti}@deis.unical.it

This work deals with the problem of classifying uncertain data. With this aim the Uncertain Nearest Neighbor (UNN) rule is here introduced, which represents the generalization of the deterministic nearest neighbor rule to the case in which uncertain objects are available. The UNN rule relies on the concept of nearest neighbor class, rather than on that of nearest neighbor object. The nearest neighbor class of a test object is the class that maximizes the probability of providing its nearest neighbor. It is provided evidence that the former concept is much more powerful than the latter one in the presence of uncertainty, in that it correctly models the right semantics of the nearest neighbor decision rule when applied to the uncertain scenario. An effective and efficient algorithm to perform uncertain nearest neighbor classification of a generic (un)certain test object is designed, based on properties that greatly reduce the temporal cost associated with nearest neighbor class probability computation. Experimental results are presented, showing that the UNN rule is effective and efficient in classifying uncertain data.

## 1. INTRODUCTION

Classification is one of the basic tasks in data mining and machine learning [Tan et al. 2005; Mitchell 1997]. Given a set of *examples* or *training set*, that is a set of objects $x_i$ with associated class labels $l(x_i)$, the goal of classification is to exploit the training set in order to build a *classifier* for prediction purposes, that is a function mapping unseen objects to one of the predefined class labels. Traditional classification techniques deal with feature vectors having *deterministic* values. Thus, data *uncertainty* is usually ignored in the learning problem formulation. However, it must be noted that uncertainty arises in real data in many ways, since the data may contain errors or may be only partially complete [Lindley 2006].

The uncertainty may result from the limitations of the equipment, indeed physical devices are often imprecise due to *measurement errors*. Another source of uncertainty are *repeated measurements*, e.g. sea surface temperature could be recorded multiple times during a day. Also, in some applications data values are *continuously changing*, as positions of mobile devices or observations associated with natural phenomena, and these quantities can be approximated by using an uncertain model.

Simply disregarding uncertainty may led to less accurate conclusions or even inexact ones. This has created the need for uncertain data management techniques [Aggarwal and Yu 2009], that are techniques managing data records typically represented by probability distributions ([Bi and Zhang 2004; Achtert et al. 2005; Kriegel and Pfeifle 2005; Ngai et al. 2006; Aggarwal and Yu 2008] to cite a few).

This work deals with the problem of classifying uncertain data. Specifically, here it is assumed that an *uncertain object* is *an object whose actual value is modeled by a multivariate probability density function*. This notion of uncertain object has been extensively adopted in the literature and corresponds to the *attribute level uncertainty model* viewpoint [Green and Tannen 2006].

Classification methods often rely on the use of distances or similarity metrics in order to implement their decision rule. It must be noted that different concepts of similarity between uncertain objects have been proposed in the literature, among them the *distance between means*, the *expected distance*, and *probabilistic threshold distance* [Lukaszyk 2004; Cheng et al. 2004; Tao et al. 2007; Agarwal et al. 2009; Angiulli and Fassetti 2011]. Thus, a seemingly suitable strategy to classify uncertain data is to make use of ad-hoc similarity metrics in order to apply to such kind of data classification techniques already designed for the deterministic setting. We call this strategy the *naive approach*.

However, in this work we provide evidence that the above depicted approach is too weak, since *there is no guarantee on the quality of the class returned by the naive approach*. As a matter of fact, *the naive approach may return the wrong class even if the probability for the object to belong to that class approaches to zero*. Hence, as a major contribution, we provide a novel classification rule which directly builds on certain similarity metrics, rather than directly exploiting ad-hoc uncertain metrics, but anyway implements a decision rule which is suitable for classifying uncertain data.

Specifically, we conduct our investigation in the context of the *Nearest Neighbor rule* [Cover and Hart 1967; Devroye et al. 1996], since it allows to directly exploit similarity metrics to the classification task. The nearest neighbor rule assigns to an unclassified object the label of the nearest of a set of previously classified objects, and can be generalized to the case in which the $k$ nearest neighbors are taken into account [Fukunaga and Hostetler 1975]. Despite its seemingly simplicity, it is very effective in classifying data [Stone 1977; Devroye 1981; Wu et al. 2008].

As already pointed out, as the main contribution of this work a novel classification rule for the uncertain setting is introduced, called the *Uncertain Nearest Neighbor* (UNN, for short). The uncertain nearest neighbor rule relies on the concept of *nearest neighbor class*, rather than on that of nearest neighbor object, the latter concept being the one the naive approach implemented through the use of the nearest neighbor rule relies on. Consider the binary classification problem with class labels $c$ and $c'$: $c$ ($c'$, resp.) is the nearest neighbor class of the test object $q$ if the probability that the nearest neighbor of $q$ comes from class $c$ ($c'$, resp.) is greater than the probability that it comes from the other class. Such a probability *takes simultaneously into account the distribution functions of all the distances separating $q$ by the training set objects*.

Summarizing, the contributions of the work are those reported in the following:

—the concept of *nearest neighbor class* is introduced and it is shown to be much more powerful than the concept of nearest neighbor in presence of uncertainty;

—based on the concept of nearest neighbor class, the *Uncertain Nearest Neighbor classification rule* (UNN) is defined. Specifically, it is precisely shown that UNN represents the generalization of the certain nearest neighbor rule to the case in which uncertain objects, represented by means of arbitrary probability density functions, are taken into account.

—it is show than the UNN rule represents a viable way to compute the most probable class of the test object, since *properties to efficiently compute the nearest neighbor class probability* are presented;

—based on these properties, an *effective algorithm to perform uncertain nearest neighbor classification* of a generic (un)certain test object is designed.

—the *experimental campaign* confirms the superiority of the UNN rule with respect to classical classification techniques in presence of uncertainty and with respect to density based classification methods specifically designed for uncertain data. Moreover, the meaningfulness of UNN classification is illustrated through a real-life prediction scenario involving wireless mobile devices.

The rest of the paper is organized as follows. Section 2 introduces the uncertain nearest neighbor classification rule. In Section 3 the properties of the uncertain nearest neighbor rule are stated and an efficient algorithm solving the task at hand is described. Section 4 discusses relationship with related works. Section 5 reports experimental results. Finally, Section 6 draws the conclusions.

## 2.  UNCERTAIN NEAREST NEIGHBOR CLASSIFICATION

In this section the Uncertain Nearest Neighbor rule is introduced. The section is organized as follows. First, uncertain objects are formalized (Section 2.1), then the behavior of the nearest neighbor rule in presence of uncertain objects is analyzed (Section 2.2) and, finally, the uncertain nearest neighbor rule is introduced (Section 2.3).

### 2.1  Uncertain objects

Let $(\mathbb{D}, \mathrm{d})$ denote a metric space, where $\mathbb{D}$ is a set, also called *domain*, and d is a *distance metric* on $\mathbb{D}$ (e.g., $\mathbb{D}$ is the $d$-dimensional real space $\mathbb{R}^d$ equipped with the Euclidean distance d).

A *certain object* $v$ is an element of $\mathbb{D}$. An *uncertain object* $x$ is a random variable having domain $\mathbb{D}$ with associated probability density function $f^x$, where $f^x(v)$ denotes the probability for $x$ to assume value $v$. A certain object $v$ can be regarded as an uncertain one whose associated pdf $f^v$ is $\delta_v(t)$, where $\delta_v(t) = \delta(0)$, for $t = v$, and $\delta_v(t) = 0$, otherwise, with $\delta(t)$ denoting the Dirac delta function.

Given two uncertain objects $x$ and $y$, $\mathrm{d}(x, y)$ denotes the continuous random variable representing the *distance* between $x$ and $y$.

Given a set $S = \{x_1, \ldots, x_n\}$ of uncertain objects, an *outcome* $I_S$ of $S$ is a set $\{v_1, \ldots, v_n\}$ of certain objects such that $f^{x_i}(v_i) > 0$ $(1 \leq i \leq n)$. The probability $Pr(I_S)$ of the outcome $I_S$ is

$$Pr(I_S) = \prod_{i=1}^{n} f^{x_i}(v_i).$$

Given an object $v$ of $\mathbb{D}$, $\mathcal{B}_R(v)$ denotes the set of values $\{w \in \mathbb{D} \mid d(w, v) \leq R\}$, namely the *hyperball* having center $v$ and radius $R$.

### 2.2 The nearest neighbor rule in presence of uncertain objects

In this section the classic Nearest Neighbor rule is recalled and, furthermore, it is shown that its direct application to the classification of uncertain data is misleading. Hence, the concept of nearest neighbor class is introduced, which captures the right semantics of the nearest neighbor rule when applied to objects modeled by means of arbitrary probability density functions. The nearest neighbor class forms the basis upon the novel Uncertain Nearest Neighbor classification rule is built on.

*Nearest Neighbor classification rule.* Let $v$ be an (un)certain object. The class label associated with $v$ is denoted by $l(v)$.

Given a set of certain objects $T'$ and a certain object $v$, the *nearest neighbor* $nn_{T'}(v)$ of $v$ in $T'$ is the object $u$ of $T'$ such that for any other object $w$ of $T'$ it holds that $\mathrm{d}(v, u) \leq \mathrm{d}(v, w)$ (ties are arbitrarily broken).

The *$k$-th nearest neighbor* $nn_{T'}^k(v)$ of $v$ in $T'$ is the object $u$ of $T'$ such that there exist exactly $k - 1$ other objects $w$ of $T'$ for which it holds that $\mathrm{d}(v, w) \leq \mathrm{d}(v, u)$ (also in this case, ties are arbitrarily broken).

In the following, $q$ denotes a generic certain test object.

Given a labelled set of certain objects $T'$, the (certain) *Nearest Neighbor* rule $NN_{T'}(q)$ [Cover and Hart 1967] assigns to the certain test object $q$ the label of its nearest neighbor in $T'$, that is

$$NN_{T'}(q) = l(nn_{T'}(q)).$$

The nearest neighbor rule can be generalized to take into account the $k$ nearest neighbors of the test object $q$: The (certain) *$k$ Nearest Neighbor* rule $NN_{T'}^k(q)$ [Fukunaga and Hostetler 1975; Devroye et al. 1996] (or, simply, $NN_{T'}(q)$, whenever the value of $k$ is clear from the context) assigns the object $q$ to the class with the most members present among its $k$ nearest neighbors in the training set $T'$.

*Applying the Nearest Neighbor rule to uncertain data.* In order to be applied, the nearest neighbor rule merely requires the availability of a distance function. In the context of uncertain data, different similarity measures have been defined, among them the *distance between means*, representing the distance between the expected values of the two uncertain objects, and the *expected distance* [Lukaszyk 2004], representing the mean of distances between all the outcomes of the two uncertain objects.

Thus, a seemingly faithful strategy to correctly classify uncertain data is to directly exploit the nearest neighbor rule in order to determine the training set object $y$ most similar to the test object $q$ and then to return the class label $l(y)$ of $y$, also referred to as *naive approach* in the following. However, it is pointed out here that *there is no guarantee on the quality of the class returned by the naive approach*. Specifically, this approach is defective since *it can return the wrong class even if its probability approaches to zero*. Next an illustrative example it is discussed.

EXAMPLE 2.1. *Consider Figure 1(a), reporting four 2-dimensional uncertain training set objects whose support is delimited by circles/ellipsis. The certain test object $q$ is located in $(0, 0)$. The blue class consists of one normally distributed uncertain object (centered in $(0, 4)$), while the red class consists of three uncertain objects, all having bimodal distribution. To ease computations, probability values*

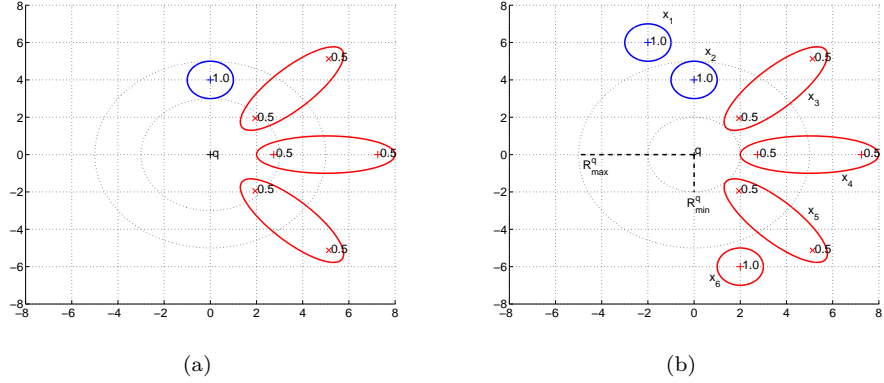

Fig. 1.  Example of comparison between the nearest neighbor object and class.

*are concentrated in the points identified by crosses.*

*It can be noticed that the object closest to $q$ according to the naive approach is that belonging to the blue class. However, it appears that the probability that a red object is closer to $q$ than a blue one is $1 - 0.5^3 = 0.875$. Thus, in the 87.5% of the outcomes of this training set the nearest neighbor of $q$ comes from the red class, but the naive approach outputs the opposite one! Note that the probability of the blue class can be made arbitrarily small by adding other red objects similar to those already present. With $n$ red objects, the probability $Pr(D(q, red) < D(p, blue))$ is $1 - 0.5^n$, that rapidly approaches to 1.*

The poor performance of the nearest neighbor rule can be explained by noticing that it takes into account the occurrence probabilities of the training set objects one at a time, a meaningless strategy in presence of many objects whose outcome is uncertain. In the following the concept of most probable class is introduced, which takes simultaneously into account the distribution functions of all the distances separating the test object by the training set objects.

*Most probable class.* Let $T = \{x_1, \ldots, x_n\}$ denote a labelled training set of uncertain objects. The probability $Pr(NN_T(q) = c)$ that the object $q$ will be assigned to class $c$ by means of the nearest neighbor rule can be computed as:

$$Pr(NN_T(q) = c) = \int_{\mathbb{D}^n} Pr(I_T) \cdot \mathbf{I}_c(NN_{I_T}(q)) \; \mathrm{d}I_T, \qquad (1)$$

where the function $\mathbf{I}_c(\cdot)$ outputs 1 if its argument equals $c$, and 0 otherwise. Informally speaking, the probability that the nearest neighbor class of $q$ in $T$ is $c$, is the summation of the occurrence probabilities of all the outcomes $I_T$ of the training set $T$ for which the nearest neighbor object of $q$ in $I_T$ has class $c$.

Thus, when uncertain objects are taken into account the nearest neighbor decision rule should output the *most probable class* $c^*$ of $q$, that is the class $c^*$ such that

$$c^* = \arg\max_c Pr(NN_T(q) = c). \qquad (2)$$

For $u$ an uncertain test object, Equation (1) becomes:

$$Pr(NN_T(u) = c) = \int_{\mathbb{D}^{n+1}} f^u(q) \cdot Pr(I_T) \cdot \mathbf{I}_c(NN_{I_T}(q)) \, \mathrm{d}q \, \mathrm{d}I_T, \qquad (3)$$

that is Equation (1) extended by taking into account also the occurrence probability of the test object $q$.

It is clear from Equations (1) and (3), that in order to determine the most probable class of $q$ it is needed to compute a multi-dimensional integral (with integration domain $\mathbb{D}^n$ or $\mathbb{D}^{n+1}$), involving simultaneously all the possible outcomes of the test object and of the training set objects.

In the following section, the uncertain nearest neighbor rule is introduced, that provides an effective method for computing the most probable class of a test object according to the nearest neighbor decision rule.

### 2.3  The uncertain nearest neighbor rule

In this section the Uncertain Nearest Neighbor classification rule (UNN) is introduced. First, the concept of *distance between an object and a class* is defined, which is conducive to the definition of *nearest neighbor class* forming the basis of the *uncertain nearest neighbor rule*. Definitions, firstly introduced for $k = 1$, for the binary classification task, and for $q$ a certain test object, are readily generalized to the case $k \geq 1$, the multiclass setting, and $q$ a possibly uncertain test object, respectively. To complete the contribution, it is formally shown that the UNN rule outputs the most probable class of the test object.

*Nearest neighbor class and UNN rule.* Let $c$ be a class label and $q$ a certain object. The *distance between* (*object*) $q$ *and* (*class*) $c$, denoted by $D(q, c)$, is the random variable whose outcome is the distance between $q$ and its $k$-th training set nearest neighbor having class label $c$.

Next it is shown how the cumulative density function of $D(q, c)$ can be computed. Let us start by considering the case $k = 1$.

Let $T_c$ denote the subset of the training set composed of the objects having class label $c$, that is

$$T_c = \{x_i \in T : l(x_i) = c\}.$$

Let $p_i(R) = Pr(\mathrm{d}(q, x_i) \leq R)$ denote the cumulative density function representing the relative likelihood for the distance between $q$ and training set object $x_i$ to assume value less than or equal to $R$, that is

$$Pr(\mathrm{d}(q, x_i) \leq R) = \int_{\mathcal{B}_R(q)} f^{x_i}(v) \, \mathrm{d}v, \qquad (4)$$

where $\mathcal{B}_R(q)$ denotes the hyper-ball having radius $R$ and centered in $q$.

Then, the cumulative density function associated with $D(q, c)$ can be obtained as follows:

$$Pr(D(q, c) \leq R) = 1 - \left( \prod_{x_i \in T_c} (1 - p_i(R)) \right), \qquad (5)$$

that is one minus the probability that no object of the class $c$ lies within distance $R$ from $q$.

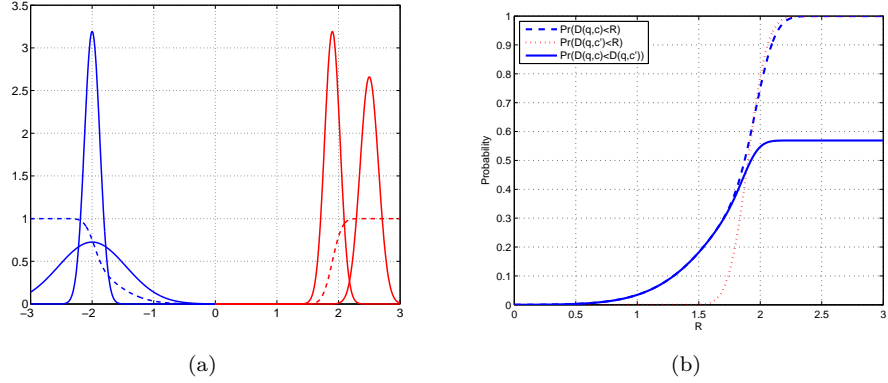

Fig. 2.   Example of distances between an object and two classes.

EXAMPLE 2.2.  *Figure 2(a) shows a one-dimensional example training set composed of four uncertain objects. The abscissa reports the domain values, while the ordinata reports the pdf values associated with the (normally distributed) uncertain objects. From left to right, means $\mu_i$ and standard deviations $\sigma_i$ are $\mu_1 = -2$ and $\sigma_1 = 0.125$, $\mu_2 = -2$ and $\sigma_2 = 0.55$, $\mu_3 = 1.9$ and $\sigma_3 = 0.125$, and $\mu_4 = 2.5$ and $\sigma_4 = 0.15$. The two objects on the left belong to class $c$ (blue), while the two other objects belong to class $c'$ (red). Consider the certain test object $q = 0$. The dashed blue (dotted red, resp.) curve represents the probability $Pr(D(q, c) \leq R)$ ($Pr(D(q, c') \leq R)$, resp.), where $R = |v|$ and $v$ denotes the abscissa value.*

Consider the *binary classification problem*, in which there are exactly two classes, with labels $c$ and $c'$, respectively. The *nearest neighbor class* of $q$ is $c$ if

$$Pr(D(q, c) < D(q, c')) \geq 0.5 \qquad (6)$$

holds, and $c'$ otherwise. In the following, the probability $Pr(D(q, c) < D(q, c'))$ is also referred to as *nearest neighbor probability of class $c$ (w.r.t. class $c'$).*

According to Equation (6), class $c$ is the nearest neighbor class of $q$ if the probability that the nearest neighbor of $q$ comes from class $c$ is greater than the probability that it comes from class $c'$.

In particular, the above probability can be computed by means of the following one-dimensional integral:

$$Pr(D(q, c) < D(q, c')) = \int_0^{+\infty} Pr(D(q, c) = R) \cdot Pr(D(q, c') > R) \, \mathrm{d}R. \qquad (7)$$

The *Uncertain Nearest Neighbor Rule* (UNN, for short) assigns to the test object $q$ the label of its nearest neighbor class.

EXAMPLE 2.3.  *Figure 2(b) reports the probabilities $Pr(D(q, c) \leq R)$ (dashed curve) and $Pr(D(q, c') \leq R)$ (dotted curve) associated with objects in Figure 2(a), together with the value of the integral in Equation (7) computed in the interval $[0, R]$ (solid curve; for large values of $R$ this curve represents the probability $Pr(D(q, c) < D(q, c'))$). In this case the probability $Pr(D(q, c) < D(q, c'))$ is equal to $0.569$.*

*Thus, the nearest neighbor class of $q$ is $c$ (the blue class, composed of the two objects on the left).*

Other the returning the nearest neighbor class $c$ of $q$, the UNN rule is also able to output the probability $p = Pr(D(q, c) < D(q, c'))$ ($p' = 1 - p$, resp.) that object $q$ belongs to class $c$ ($c'$, resp.), and it is worth to notice that the naive approach cannot provide such a value.

Moreover, it is also important to point out that enabling the nearest neighbor rule to handle uncertain data makes it more robust to noise with respect to the case in which the uncertainty associated with data is ignored.

EXAMPLE 2.4. *As an example, consider the uncertain objects in Figure 1(a) and assume to disregard uncertainty by replacing them with their means. In such a case, the certain nearest neighbor rule will erroneously output the blue label, since the noisy blue object (centered in $(4, 0)$) is closer to $q$ than the red ones. Contrarily, the uncertain nearest neighbor correctly classifies $q$, since it simultaneously considers the whole class distribution.*

Informally speaking, it can be said the distribution of the closest class tends to overshadow noisy objects.

*Generalizing the UNN rule.* The Uncertain Nearest Neighbor rule can be readily generalized in order $(a)$ to take into account arbitrary values of $k$, $(b)$ to consider possibly uncertain test objects and $(c)$ to deal with the multiclass problem, as accounted for in the following.

$(a)$ The first point can be achieved by properly redefining the probability $Pr(D(q, c) \leq R)$. For $k \geq 1$, the probability $Pr(D(q, c) \leq R)$ can be expressed as follows:

$$1 - \left( \sum_{S \subseteq T_c : |S| < k} \left( \left( \prod_{x_i \in S} p_i(R) \right) \cdot \left( \prod_{x_i \in (T_c \setminus S)} (1 - p_i(R)) \right) \right) \right), \qquad (8)$$

that is one minus the probability of having less than $k$ objects of class $c$ lying within distance $R$ from $q$.

In particular, for $k = 1$, the summation involves only the empty set $S = \emptyset$ (which is the unique subset of $T_c$ having size smaller than one), and the expression reduces to $1 - (\prod_{x_i \in T_c} (1 - p_i(R)))$, that is the expression already reported in Equation (5).

$(b)$ Assume now to have an uncertain test object $u$. In this case, it holds that

$$Pr(D(u, c) < D(u, c')) = \int_{\mathbb{D}} f^u(q) \cdot Pr(D(q, c) < D(q, c')) \, \mathrm{d}q =$$
$$= \int_{\mathbb{D}} \int_0^{+\infty} f^u(q) \cdot Pr(D(q, c) = R) \cdot Pr(D(q, c') > R) \, \mathrm{d}R \, \mathrm{d}q, \qquad (9)$$

and the nearest neighbor class of $q$ is defined as that class $c$ such that $Pr(D(u, c) < D(u, c')) \geq 0.5$.

$(c)$ Till now only the binary classification problem has been taken into account. In order to deal with the *multiclass classification problem*, the *one-against-all*-like [Rifkin and Klautau 2004] approach is adopted. Assume there are $m$ classes, with labels $c_1, c_2, \ldots, c_m$. For each class $c_j$, let $\bar{c}_j$ denote a novel class representing the union of all classes except for $c_j$. Then object $x_i$ belongs to class $\bar{c}_j$ (and, hence, $l(x_i) = \bar{c}_j$) if $l(x_i) \neq c_j$ $(1 \leq j \leq m)$. The *(uncertain) nearest neighbor class* of the (un)certain test object $q$ is

$$\arg\max_{c_j} Pr\left(D(q, c_j) < D(q, \bar{c}_j)\right),\qquad(10)$$

that is the class that maximizes the probability to provide the $k$-nearest neighbors of the test object.

*Equivalence with the most probable class.* One of the main properties of the UNN rule is now stated. Indeed, the following theorem formally proves that the uncertain nearest neighbor rule captures the right semantics of the nearest neighbor rule when uncertain data is taken into account.

THEOREM 2.5. *The Uncertain Nearest Neighbor rule outputs the most probable class (see Equation (2)) of the test object.*

In order to prove the above statement, first it is introduced the *nearest distance* decision rule and then it is shown that it relates the certain and the uncertain nearest neighbor rules.

Let $T'$ be a certain training set, with objects coming from two classes with labels $c$ and $c'$, respectively, and let $k$ be a positive integer. Then, the *nearest distance* $\mathrm{ND}_{T'}^k(q)$ is the following decision rule: output class $c$, if

$$\mathrm{d}(q, nn_{T_c'}^k(q)) < \mathrm{d}(q, nn_{T_{c'}'}^k(q)),\qquad(11)$$

and output class $c'$, otherwise.

PROPOSITION 2.6. *Consider the binary classification problem, and let $k$ be an odd positive integer. Then, it holds that*

$$\mathrm{NN}_{T'}^k(q) = \mathrm{ND}_{T'}^{\lceil \frac{k}{2} \rceil}(q).$$

PROOF. Assume that $\mathrm{NN}_{T'}^k(q)$ outputs class $c$. Then, among the $k$ nearest neighbors of $q$ in $T'$ there are at least $k' = \lceil \frac{k}{2} \rceil$ objects coming from class $c$, since the majority of the $k$ nearest neighbors have class label $c$, and less than $k'$ objects coming from class $c'$. Thus, it holds that

$$\mathrm{d}(q, nn_{T_c'}^{\lceil \frac{k}{2} \rceil}(q)) \leq \mathrm{d}(q, nn_{T'}^k(q)) < \mathrm{d}(q, nn_{T_{c'}'}^{\lceil \frac{k}{2} \rceil}(q)),$$

that is the distance separating $q$ from its $k'$-th nearest neighbor in class $c'$ (being greater than $\mathrm{d}(q, nn_{T'}^k(q))$) is greater than the distance separating $q$ from its $k'$-th nearest neighbor in class $c$ (being not greater than $\mathrm{d}(q, nn_{T'}^k(q))$), and $\mathrm{ND}_{T'}^{k'}(q)$ outputs class $c$ as well.

Vice versa, assume that $\mathrm{ND}_{T'}^{k'}(q)$ outputs class $c$. Then, the distance separating $q$ from its $k'$-th nearest neighbor in class $c$ is stricly smaller than the distance separating $q$ from its $k'$-th nearest neighbor in class $c'$. Thus, being $2k' = k + 1$, it is the case that among the $k$ nearest neighbors of $q$ in $T'$ there are at least $k'$ objects coming from class $c$ and at most $k'-1$ objects coming from class $c'$. Hence, $\mathrm{NN}_{T'}^k(q)$ outputs class $c$ as well.  □

According to the above property, it is the case that $\mathrm{NN}^1_{T'}(q) = \mathrm{ND}^1_{T'}(q)$, $\mathrm{NN}^3_{T'}(q) = \mathrm{ND}^2_{T'}(q)$, $\mathrm{NN}^5_{T'}(q) = \mathrm{ND}^3_{T'}(q)$, $\mathrm{NN}^7_{T'}(q) = \mathrm{ND}^4_{T'}(q)$, and so on.

Now, the proof of Theorem 2.5 can be given.

PROOF OF THEOREM 2.5. The proof follows from Proposition 2.6 and Equations (6) and (11), by noticing that $D(q, c)$ is the random variable representing the distance between the certain test object $q$ and its $k$-th nearest neighbor in the class $c$ of the uncertain dataset. $\square$

## 3. CLASSIFYING A TEST OBJECT

This section presents then UNN classification algorithm. First, some preliminary definitions are provided (Section 3.1), then properties of the nearest neighbor class probability are stated (Section 3.2) and it is shown how this probability can be computed (Section 3.3) and, finally, the UNN classification algorithm is described (Section 3.4). The last Section 3.5, discusses how some steps of the UNN algorithm can be accelerated in practice.

### 3.1 Preliminaries

Without loss of generality it is assumed that each uncertain object $x$ is associated with a finite region $\mathrm{SUP}(x)$, containing the support of $x$, namely the region such that $Pr(x \notin \mathrm{SUP}(x)) = 0$ holds. If the support of $x$ is infinite, then $\mathrm{SUP}(x)$ is such that $Pr(x \notin \mathrm{SUP}(x)) \leq \alpha$, for a fixed small value $\alpha$, and the probability for $x$ to exist outside $\mathrm{SUP}(x)$ is considered negligible.

It must be noticed that these assumptions are not restrictive, since the error $\epsilon = \alpha^2$ involved in the calculation of the probability $Pr(d(x, y) \leq R)$, with $x$ and $y$ two uncertain objects, can be made arbitrarily small by properly selecting the regions $\mathrm{SUP}(x)$ and $\mathrm{SUP}(y)$ and, hence, the value $\alpha$.

For example, assume that the data set objects $x \in \mathbb{R}^d$ are normally distributed with mean $\mu_j$ and standard deviation $\sigma_j$ along each dimension $j$ ($1 \leq j \leq d$). If the region $\mathrm{SUP}(x)$ is defined as $[\mu_j - 4\sigma_j, \mu_j + 4\sigma_j]^d$ then the probability $\alpha = Pr(x \notin \mathrm{SUP}(x))$ is $\alpha = (2 \cdot \Phi(-4))^d \approx 0.00006^d$ and the maximum error is $\epsilon = \alpha^2 \approx (4 \cdot 10^{-9})^d$.

The region $\mathrm{SUP}(x)$ can be defined as an hyperball or an hypercube. In the former case, $\mathrm{SUP}(x)$ is identified by means of its center $c(x)$ and its radius $r(x)$, where $c(x)$ is a certain object and $r(x)$ is a positive real number.

The *minimum distance $mindist(x, y)$* between $x$ and $y$ is defined as

$$\min\{\mathrm{d}(v, w) : v \in \mathrm{SUP}(x) \wedge w \in \mathrm{SUP}(y)\} = \max\{0, d(c(x), c(y)) - r(x) - r(y)\},$$

while the *maximum distance $maxdist(x, y)$* between $x$ and $y$ is defined as

$$\max\{\mathrm{d}(v, w) : v \in \mathrm{SUP}(x) \wedge w \in \mathrm{SUP}(y)\} = d(c(x), c(y)) + r(x) + r(y).$$

Given a set of objects $S$, let $inner_S(q, R)$ denote the set $\{x \in S : maxdist(q, x) \leq R\}$, that is the subset of $S$ composed of all the objects $x$ whose maximum distance from $q$ is not greater than $R$.

Let $R^q_c$ denote the positive real number

$$R^q_c = \min\{R \geq 0 : |inner_{T_c}(q, R)| \geq k\}, \tag{12}$$

representing the smallest radius $R$ for which there exist at least $k$ objects of the class $c$ having maximun distance from $q$ not greater than $R$.

Moreover, let $R_{max}^q$ be defined as

$$R_{max}^q = \min\{R_c^q, R_{c'}^q\}, \tag{13}$$

and let $R_{min}^q$ be defined as

$$R_{min}^q = \min_{x \in T} mindist(q, x). \tag{14}$$

## 3.2    Properties

This section presents two important properties of the nearest neighbor class probability. In particular, it is shown that in order to compute the probability reported in Equation (7) a specific finite domain can be considered instead of $\mathbb{R}_0^+$ (Proposition 3.1) and also a specific subset of the training set of objects can be taken into account instead of the whole training set (Proposition 3.3).

These properties have direct practical implications, since they allow the computation of the nearest neighbor class probability by means of a less demanding integration formula (see Equation (15) at the end of the section).

Let us start by considering the integration domain.

PROPOSITION 3.1. *It holds that*

$$Pr(D(q, c) < D(q, c')) = \int_{R_{min}^q}^{R_{max}^q} Pr(D(q, c) = R) \cdot Pr(D(q, c') > R) \, \mathrm{d}R.$$

PROOF. Assume that $R_{max}^q$ be $R_{c'}^q$, then there exist (at least) $k$ objects $x_{j_1}$, ..., $x_{j_k}$ of the class $c'$ such that $maxdist(q, x_{j_i}) \le R_{max}^q$ $(1 \le i \le k)$. For $R > R_{max}^q$, it holds that $p_{j_i}(R) = 1$, since the support $\mathrm{SUP}(x_{j_i})$ of $x_{j_i}$ is within distance $R$ from $q$. Thus, in Equation (5) the summation over all subsets $S$ of $T_c$ having size strictly less than $k$ evaluates to zero, and $Pr(D(q, c') \le R) = 1$. Indeed, for each subset $S$ of $T_c$, there exists at least one object $x_{j_h}$ in the set $\{x_{j_1}, \ldots, x_{j_k}\}$ which is not in $S$, and, hence, at least one term $(1 - p_{j_h}(R)) = 1 - 1 = 0$ in the productory. As a consequence, for each $R > R_{max}^q$, the term $Pr(D(q, c') > R) = 1 - Pr(D(q, c') \le R)$ in the integral of Equation (7) is null, and the computation of the integral can be restricted to the interval $[0, R_{max}^q]$.

Conversely, assume that $l(x^q) = c$. By adopting a very similar line of reasoning, it can be concluded that for each $R > R_{max}^q$ the probability $Pr(D(q, c) = R)$ is null.

Since for $R < R_{min}^q$, $Pr(D(q, c) < D(q, c'))$ is zero, the result follows.    □

From the practical point of view, the above property has the important implication that in order to determine the probability $Pr(D(q, c) < D(q, c'))$, it suffices to compute the integral reported in Equation (7) on the finite domain $[R_{min}^q, R_{max}^q]$.

EXAMPLE 3.2. *Consider Figure 1(b). For $k = 1$, the value $R_{max}^q$ denotes the radius of the smallest hyperball centered in $q$ that entirely contains the support of one training set object, hence it is equal to $maxdist(q, x_2)$. The value $R_{min}^q$ denotes the radius of the greatest hyperball centered in $q$ that does not contain the support of any training set object, hence it is equal to $mindist(q, x_3)$.*

PROPOSITION 3.3. *Let $T^q$ be the set composed of the training set objects $x_i$ such that $mindist(q, x_i) \leq R^q_{max}$, and let $D_q(c)$ the random variable whose outcome is the distance between $q$ and its $k$-th nearest neighbor in the set $T^q$ having class label $c$. Then, it holds that $Pr(D(q, c) < D(q, c')) = Pr(D_q(c) < D_q(c'))$.*

PROOF. In order to prove the property it suffices to show that the training set objects $x_i$ such that $mindist(q, x_i) > R^q_{max}$ do not contribute to the computation of the probability $Pr(D(q, c) < D(q, c'))$.

Assume that $R \leq R^q_{max}$, let $x_j$ be a generic object such that $mindist(q, x_j) > R^q_{max}$, and consider the subset $T'_c = T_c \setminus \{x_j\}$ of $T_c$. Let $n$ be the number of objects in $T'_c$. Now it is shown that the value of the probability $Pr(D(q, c) \leq R)$ computed on the sets $T'_c$ and $T_c$ is identical. Consider the summation in Equation (8) over all the subsets $S'$ of $T'_c$ having size $|S'|$ less than $k$. The value of the same summation over all the subsets $S$ of $T_c$ having size $|S|$ less than $k$ can be obtained by considering the following number of terms:

$$\sum_{\ell=0}^{k-1} \binom{n+1}{\ell} = 1 + \sum_{\ell=1}^{k-1} \left[ \binom{n}{\ell-1} + \binom{n}{\ell} \right] = \binom{n}{k-1} + 2 \sum_{\ell=0}^{k-2} \binom{n}{\ell}.$$

That is to say, with each term $t$ in the summation over $T'_c$, concerning the subset $S'$ of $T'_c$ having less than $k-1$ elements (exactly $k-1$ elements, resp.), two terms $t'$ and $t''$ are associated with (one term $t'$ is associated with, resp.) in the summation over $T_c$. In particular, $t'$ concerns the subset $S = S'$ and $t''$ concerns the subset $S = S' \cup \{x_j\}$. As for the terms $t'$, since $x_j \notin S'$, it holds that $t' = t \cdot (1 - p_j(R)) = t \cdot (1 - 0) = t$, since $p_j(R) = 0$ for each $R \leq R^q_{max}$ (recall that $mindist(q, x_j) > R^q_{max}$). As for the terms $t''$, since $x_j \in S''$, it then holds that $t'' = t \cdot p_j(R) = t \cdot 0 = 0$. It can be concluded that the two summations coincide and, hence, that all objects $x_j$ can be safely ignored.

As for $R > R^q_{max}$, the result follows from Property 3.1.  □

Also the above property has an important practical implication. Indeed, it states that, once the test object $q$ is given, in order to determine the probability $Pr(D(q, c) < D(q, c'))$, the computation can be restricted to the set $T^q$ composed of the training set objects $x_i$ such that $mindist(q, x_i) \leq R^q_{max}$.

EXAMPLE 3.4. *Consider again the example of Figure 1(b). Then, the set $T^q$ consists of the objects $x_2$, $x_3$, $x_4$, and $x_5$, and objects $x_1$ and $x_6$ do not contribute to the computation of the integral in Equation (7).*

By putting things together, the following result can be eventually obtained.

THEOREM 3.5. *For any (un)certain test object $q$, it holds that*

$$Pr(D(q, c) < D(q, c')) = \int_{R^q_{min}}^{R^q_{max}} Pr(D_q(c) = R) \cdot Pr(D_q(c') > R) \, \mathrm{d}R. \qquad (15)$$

PROOF. The result follows from Propositions 3.1 and 3.3.  □

### 3.3 Computing the nearest neighbor class probability

In this section it is shown how the value of the integral in Equation (15) can be obtained. This integral depends on probabilities $Pr(D_q(c) = R)$ and $Pr(D_q(c') > R)$, which in their turn depend on probabilities $p_i(R)$. Moreover, functions $p_i(R)$ depend on the objects $x_i$ and $q$ and, for any given value of $R$, they involve the computation of one multi-dimensional integral with domain of integration the hyper-ball in $\mathbb{D}$ of center $q$ and radius $R$.

Next methods to compute as efficiently as possible probabilities $p_i(R)$ (Section 3.3.1), the class distance probability (Section 3.3.2), and the nearest neighbor class (Section 3.3.3) are described.

3.3.1 *Computation of the probabilities $p_i(R)$.* Next, it is considered the most general case of arbitrarily shaped multi-dimensional pdfs, having as domain $\mathbb{D}$ the $d$-dimensional Euclidean space $\mathbb{R}^d$. It is known [Lepage 1978] that given a function $g$, if $N$ points $w_1, w_2, \ldots, w_N$ are randomly selected according to a given pdf $f$, then the following approximation holds

$$\int g(v)\,\mathrm{d}v \approx \frac{1}{N} \sum_{j=1}^{N} \frac{g(w_j)}{f(w_j)}. \tag{16}$$

Thus, in order to compute the value $p_i(R)$, the function $g_i(v) = f^{x_i}(v)$ if $d(q,v) \leq R$, and $g_i(v) = 0$ otherwise, can be integrated by evaluating formula in Equation (16) with the points $w_j$ randomly selected according to the pdf $f^{x_i}$. This procedure reduces to compute the relative number of sample points $w_j$ lying at distance not greater than $R$ from $q$, that is

$$p_i(R) = \frac{|\{w_j : d(q, w_j) \leq R\}|}{N}.$$

More precisely, by exploiting this kind of strategy a suitable approximation of *the whole cumulative distribution function $p_i$ can be computed with only one single integration* operation, as shown in the following.

With each function $p_i$ an histogram $H_i$ of $h$ slots (with $h$ a parameter used to set the resolution of the histogram) representing the value of the function $p_i$ in the interval $[R^q_{min}, R^q_{max}]$ is associated. Let $\Delta R$ be

$$\Delta R = \frac{(R^q_{max} - R^q_{min})}{h}$$

and $R_l$ be

$$R_l = R^q_{min} + l \cdot \Delta R,$$

then the $l$th slot $H_i(l)$ of $H_i$ stores the value $p_i(R_l)$ $(1 \leq l \leq h)$. After having generated the $N$ points $w_1, w_2, \ldots, w_N$ according to the pdf $f^{x_i}$, each entry $H_i(l)$ can be eventually obtained as

$$H_i(l) = \frac{|\{w_j : d(q, w_j) \leq R_l\}|}{N},$$

where distances $d(q, w_j)$ are computed once and reused during the computation of each slot value.

3.3.2 *Class distance probability computation.* In this section we show how the probability $Pr(D_q(c) \leq R)$ of having at least $k$ objects within distance $R$ from $q$, can be computed.

Assume that an arbitrary order among the elements of the set $T_c^q$ is given, namely $T_c^q = \{x_1, \ldots, x_{|T_c^q|}\}$. Then, the probability $D_q(c)$ corresponds to the probability that an element of $T_c^q$ is actually the $k$-th object (according to the established order) lying within distance $R$ from $q$.

By letting $P_c^q(i, j)$ denote the probability that exactly $i$ objects among the first $j$ objects of $T_c^q$ lie within distance $R$ from $q$ ($i \geq 0, j \leq |T_c^q|$), it follows that

$$p_j(R) \cdot P_c^q(k-1, j-1)$$

represents the probability that the $j$-th element of $T_c^q$ lies within distance $R$ from $q$ and exactly $k-1$ objects preceding $x_j$ in $T_c^q$ lie within distance $R$ from $q$.

Thus, the probability $Pr(D_q(c) \leq R)$ can be rewritten as

$$Pr(D_q(c) \leq R) = \sum_{1 \leq j \leq |T_c^q|} p_j(R) \cdot P_c^q(k-1, j-1). \qquad (17)$$

The probability $P_c^q(i, j)$ can be recursively computed as follows:

$$P_c^q(i, j) = p_j(R) \cdot P_c^q(i-1, j-1) + (1 - p_j(R)) \cdot P_c^q(i, j-1). \qquad (18)$$

Indeed, the probability $P_c^q(i, j)$ corresponds to the probability that $x_j$ lies within distance $R$ from $q$ and exactly $i-1$ objects among the first $j-1$ objects of $T_c$ lie within distance $R$ from $q$, plus the probability that $x_j$ does not lie within distance $R$ from $q$ and exactly $i$ objects among the first $j-1$ objects of $T_c$ lie within distance $R$ from $q$.

As for the properties of $P_c^q(i, j)$, we note that

1. $P_c^q(0, 0) = 1$: since it corresponds to the probability that exactly 0 objects among the first 0 objects of $T_c^q$ lie within distance $R$ from $q$;
2. $P_c^q(0, j) = \Pi_{1 \leq h \leq j}(1 - p_h(R))$, with $j > 0$: since it corresponds to the probability that none of the first $j$ objects of $T_c^q$ lie within distance $R$ from $q$;
3. $P_c^q(i, j) = 0$ with $i > j$: since if $j < i$ it is not possible that $i$ objects among the first $j$ objects of $T_c^q$ lie within distance $R$ from $q$.

Technically, the probability $P_c^q(i, j)$ can be computed by means of a dynamic programming procedure, similarly to what shown in [Rushdi and Al-Qasimi 1994]. The procedure makes use of a $k \times (|T_c^q| + 1)$ matrix $M_c^q$: The generic element $M_c^q(i, j)$ stores the the probability $P_c^q(i, j)$. Due to property 3 above, $M_c^q$ is an upper triangular matrix, namely all the elements below the main diagonal are equal to 0. The first row of $M_c^q$ is computed by applying properties 1 and 2 above. Then, the procedure fills the matrix $M_c^q$ (from the second to the $k$-th row) by applying Equation (18). The value of $D(q, c)$ is, finally, computed by exploiting the elements of the last row of $M_c^q$ in Equation (17).

As for the temporal cost required to compute Equation (17), assuming that the values $p_h(R)$ are already available ($1 \leq h \leq |T_c^q|$), from the above analysis it follows that the temporal cost is $O(k \cdot |T_c^q|)$, hence linear both in $k$ and in the size $|T_c^q|$ of $T_c^q$. As far as the spatial cost is concerned, in order to fill the $i$-th row of $M_c^q$, only the elements of the $(i-1)$-th and $i$-th rows of $M_c^q$ are required, then the procedure employs just two arrays of $|T_c^q|$ floating point numbers, and hence the space is linear in the size $|T_c^q|$ of $T_c^q$.

3.3.3 *Computation of the class probability.* In order to compute the integral reported in Equation (15), an histogram $F_c$ composed of $h$ slots is associated with the class $c$. In particular, the slot $F_c(l)$ $(1 \leq l \leq h)$ of $F_c$ stores the value $Pr(D_q(c) \leq R_l)$ computed by exploiting the procedure described in Section 3.3.2.

Then, the probability $Pr(D_q(c) = R_l)$ can be obtained as

$$\frac{Pr(D_q(c) \leq R_l) - Pr(D_q(c) \leq R_{l-1})}{\Delta R} = \frac{F_c(l) - F_c(l-1)}{\Delta R},$$

and the probability $Pr(D_q(c) < D_q(c'))$ as

$$\sum_{l=1}^{h} \left[ Pr(D_q(c) = R_l) \cdot Pr(D_q(c') > R_l) \cdot \Delta R \right].$$

To conclude, the previous summation can be finally simplified thus obtaining the following formula

$$\sum_{j=1}^{h} \left[ (F_c(l) - F_c(l-1)) \cdot (1 - F_{c'}(l)) \right], \tag{19}$$

whose value corresponds to the probability $Pr(D(q, c) < D(q, c'))$.

If the test object $u$ is uncertain, the nearest neighbor probability of class $c$ is expressed by the integral reported in Equation (9). By using formula in Equation (16) with $g(q) = f^u(q) \cdot Pr(D(q, c) < D(q, c'))$ and $f(q) = f^u(q)$ and by generating $N$ points $q_1, q_2, \ldots, q_N$ according to the pdf $f^u$, the value of the integral in Equation (9) can be obtained as

$$\frac{1}{N} \sum_{i=1}^{N} Pr(D(q_i, c) < D(q_i, c')), \tag{20}$$

where the terms $Pr(D(q_i, c) < D(q_i, c'))$ are computed by exploiting the expression in Equation (19).

## 3.4 Classification Algorithm

Figure 3 shows the *Uncertain Nearest Neighbor Classification* algorithm, which exploits properties introduced in Sections 3.2 and 3.3 in order to classify certain test objects.

The step 1 of the algorithm determines $R_{max}^q$ (see Equation (13) and Proposition 3.1), while the step 2 determines the set $T^q$ (see Proposition 3.3). As for the step 3, if one of the two classes has less than $k$ objects in $T^q$, then the object $q$ is safely assigned to the other class. Otherwise, the nearest neighbor class probability must be computed, which is accounted for in the subsequent steps by exploiting the technique described in Section 3.3.

*Temporal cost.* As far as the temporal cost of the algorithm is concerned, both steps 1 and 2 cost $O(nd)$, where $n$ is the number of training set objects and $O(d)$ is the cost of evaluating the distance between two certain objects. Let $n_q$ $(\leq n)$ be the cardinality of the set $T^q$. Step 3 costs $O(n_q)$, while step 4 costs $O(n_q d)$. As for step 5, it involves the computation of $n_q$ histograms $H_i$, each of which costs $O(Nd)$,

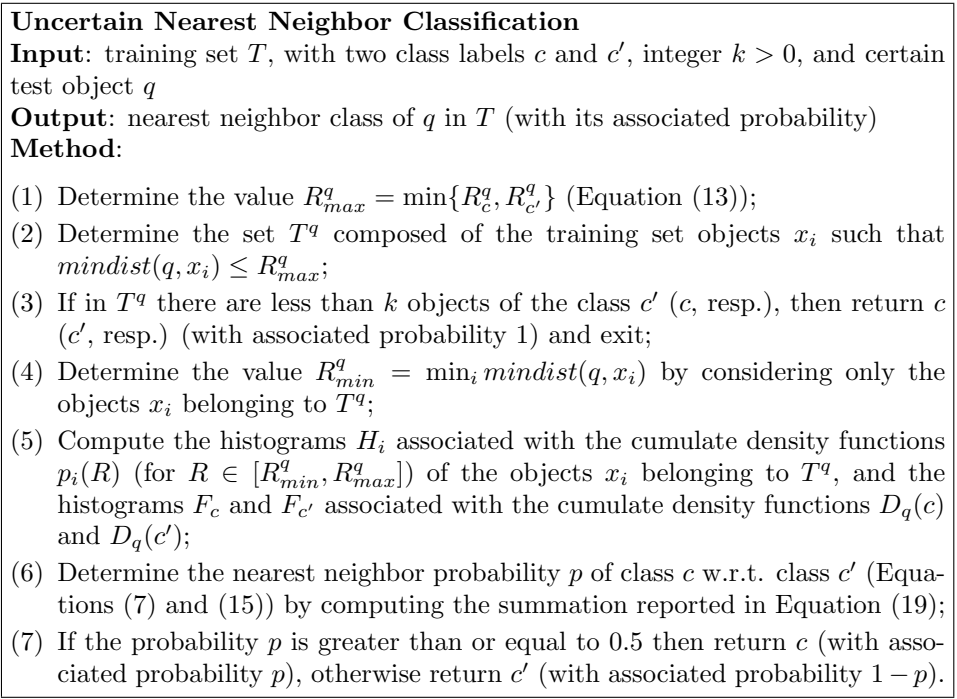

**Uncertain Nearest Neighbor Classification**

**Input**: training set $T$, with two class labels $c$ and $c'$, integer $k > 0$, and certain test object $q$

**Output**: nearest neighbor class of $q$ in $T$ (with its associated probability)

**Method**:

(1) Determine the value $R_{max}^q = \min\{R_c^q, R_{c'}^q\}$ (Equation (13));

(2) Determine the set $T^q$ composed of the training set objects $x_i$ such that $mindist(q, x_i) \leq R_{max}^q$;

(3) If in $T^q$ there are less than $k$ objects of the class $c'$ ($c$, resp.), then return $c$ ($c'$, resp.) (with associated probability 1) and exit;

(4) Determine the value $R_{min}^q = \min_i mindist(q, x_i)$ by considering only the objects $x_i$ belonging to $T^q$;

(5) Compute the histograms $H_i$ associated with the cumulate density functions $p_i(R)$ (for $R \in [R_{min}^q, R_{max}^q]$) of the objects $x_i$ belonging to $T^q$, and the histograms $F_c$ and $F_{c'}$ associated with the cumulate density functions $D_q(c)$ and $D_q(c')$;

(6) Determine the nearest neighbor probability $p$ of class $c$ w.r.t. class $c'$ (Equations (7) and (15)) by computing the summation reported in Equation (19);

(7) If the probability $p$ is greater than or equal to 0.5 then return $c$ (with associated probability $p$), otherwise return $c'$ (with associated probability $1 - p$).

Fig. 3.    The uncertain nearest neighbor classification algorithm.

with $N$ the number of points considered during integration. The computation of histograms $F_c$ and $F_{c'}$ costs $O(n_q k h)$, with $h$ the resolution of the histograms. Finally, step 6 costs $O(h)$.

It can be noticed that the term $n_q k h$ is negligible with respect to the term $n_q N d$, since $k$ is a small integer number ($k = 1$ by default, and, in any case, it is a very small integer), while it has been experimentally verified that $h = 100$ provides good quality results. Summarizing, the *temporal cost* of the algorithm is $O(n_q N d)$, with $n_q$ expected to be much smaller than $n$.

As for uncertain test objects, in order to classify them the summation in Equation (20) has to be computed. This can be accomplished by executing $N$ times the algorithm in Figure 3, with a total temporal cost $O(n_q N^2 d)$ and with no additional spatial cost.

*Spatial cost.* As far as the *spatial cost* of the algorithm is concerned, the method needs to store, other than the training set, the $n_q$ identifiers of the objects in $T^q$, the histograms $H_i$, and the two histograms $F_c$ and $F_{c'}$ consisting of $h$ floating point numbers. Summarizing, the spatial cost is $O(n_q h)$.

## 3.5 Accelerating the computation of the set $T^q$.

Before leaving the section, the computation of the set $T^q$ is discussed.

The basic strategy to compute the set $T^q$ consists in performing two linear scans of the training set objects in order to determine the radius $R_{max}^q$ (step 1 of the algorithm) and to collect objects $x_i$ such that $maxdist(q, x_i) \leq R_{max}^q$ (step 2 of the algorithm).

It can be noted that step 1 of the UNN algorithm corresponds to a *nearest neighbor query search* with respect to the value $maxdist(q, x_i)$, while step 2 corresponds to a *range query search* with radius $R_{max}^q$ with respect to the value $mindist(q, x_i)$.

Let $p$ be a certain object and $v$ and $w$ be two certain objects. Let $\delta_p(v, w)$ denote the positive real value $|\mathrm{d}(v, p) - \mathrm{d}(p, w)|$. Then, the two following relationships are satisfied:[1]

$$\delta_{p_j}(c(q), c(x_i)) + r(q) + r(x_i) \leq maxdist(q, x_i),$$

and

$$\delta_{p_j}(c(q), c(x_i)) - r(q) - r(x_i) \leq mindist(q, x_i).$$

Indeed, by the reverse triangle inequality it is the case that $\delta_{p_j}(c(q), c(x_i)) \leq d(c(q), c(x_i))$.

Thus, the two above introduced inequalities can be used as *pruning rules* to be embedded in exiting certain *similarity search methods* for metric spaces, such as pivot-based indexes, VP-trees, and others [Chávez et al. 2001], in order to fasten execution of steps 1 and 2.

It can be noticed that the above depicted strategy does not modify the asymptotic time complexity of the algorithm.

However, in practice the execution time of the algorithm can take advantage of this strategy when the cost of computing the probability $p_i(R)$ is comparable to the cost of computing the distance between the center of the test object $c(q)$ and the center of the training set object $c(x_i)$ and, moreover, the number of training set objects is very large (as an example, consider pdfs stored in histograms of fixed size).

## 4. RELATED WORK

Besides the literature concerning the classic nearest neighbor rule [Cover and Hart 1967; Stone 1977; Fukunaga and Hostetler 1975; Devroye 1981; Devroye et al. 1996], the works most related to the present one concern similarity search methods for uncertain data and classification in presence of uncertainty.

Several *similarity search methods* designed to efficiently retrieve the most similar objects of a query object have been designed [Chávez et al. 2001; Zezula et al. 2006]. These methods can be partitioned in those suitable for vector spaces [Bentley 1975; Beckmann et al. 1990; Berchtold et al. 1996], which allow to use geometric and coordinate information, and those applicable in general metric spaces [Yianilos 1993; Micó et al. 1994; Chávez et al. 2001; Zezula et al. 2006], where the above information is unavailable. The certain nearest neighbor rule may benefit of these methods since they fasten the search for the nearest neighbor of the test object. Moreover, as discussed in Section 3.5, these methods can be employed within the technique here described in order to accelerate some basic steps of the computation of the nearest neighbor class.

The above mentioned methods have been designed to be used with similarity measures involving certain data. Different concepts of similarity between uncertain objects have been proposed in the literature, among them the *distance between means*, the *expected distance*, and *probabilistic threshold distance* [Lukaszyk 2004; Cheng et al. 2004; Tao et al. 2007; Agarwal et al. 2009]. Based on some of these notions, *similarity search methods* designed to efficiently retrieve the most similar objects of a query object have been also designed. The problem of searching over uncertain data was first introduced in [Cheng et al. 2004] where the authors considered the problem of querying one-dimensional real-valued uniform pdfs. In [Ngai et al. 2006] various pruning methods to avoid the expensive expected distance calculation are introduced. Since the expected distance is a metric, the triangle inequality, involving some pre-computed expected distances between a set of anchor objects and the uncertain data set objects, can be straightforwardly employed in order to prune unfruitful distance computations. [Singh et al. 2007] considered the problem of indexing categorical uncertain data. To answer uncertain queries [Tao et al. 2007] introduced the concept of probabilistic constrained rectangles (PCR) of an object. [Angiulli and Fassetti 2011] introduced a technique to efficiently answer range queries over uncertain objects in general metric spaces.

---

[1]If $q$ is a certain object, the $c(q) = q$ and $r(q) = 0$.

While certain neighbor classification can be almost directly built on top of efficient indexing techniques for nearest neighbor search, we have already showed that the straight use of uncertain nearest neighbor search methods for classification purposes leads to a poor decision rule in the uncertain scenario. Thus, it must be pointed out that the UNN method is only loosely related to uncertain nearest neighbor indexing techniques. Moreover, as far as the efficiency of the UNN is concerned, none of these indexing methods can be straightforwardly employed to improve execution time of UNN, since they are tailored on a specific notion of similarity among uncertain objects, while UNN relies on the concept of nearest neighbor class which is directly built on a certain similarity metrics.

Recently, several mining tasks have been investigated in the context of uncertain data, including clustering, frequent pattern mining, and outlier detection [Ngai et al. 2006; Achtert et al. 2005; Kriegel and Pfeifle 2005; Aggarwal and Yu 2008; Aggarwal 2009; Aggarwal and Yu 2009]. Particularly, a few classification methods dealing with uncertain data have been proposed in the literature, among them [Mohri 2003; Bi and Zhang 2004; Aggarwal 2007]. [Mohri 2003] considered the problem of classifying uncertain data represented by means of distributions over sequences, such as weighted automata, and extended support vector machines to deal with distributions by using general kernels between weighted automata. This kind of technique is particularly suited for natural language processing applications. [Bi and Zhang 2004] investigates a learning model in which the input data is corrupted with noise. It is assumed that input objects $x_i' = x_i + \Delta x_i$ are subject to additive noise, where $x_i$ is a certain object and the noise $\Delta x_i$ follows a specific distribution. Specifically, a bounded uncertainty model is considered, that is to say $||\Delta x_i|| \leq \delta_i$ with uniform priors, and a novel formulation of support vector classification, called total support vector classification (TSVC) algorithm, is proposed to manage this kind of uncertainty. In [Aggarwal 2007] a method for handling error-prone and missing data with the use of density based approaches is presented. The estimated error associated with the $j$th dimension ($1 \leq i \leq d$) of the $d$-dimensional data point $x_i$ is denoted by $\psi_j(x_i)$. This error value may be for example the standard deviation of the observations over a large number of measurements. The basic idea of the framework is to construct an error-adjusted density of the data set by exploiting kernel density estimation and, then, to use this density as an intermediate representation in order to perform mining tasks. An algorithm for the classification problem is presented, consisting in a density based adaptation of rule-based classifiers. Intuitively, the methods seeks for the subspaces in which the instance-specific local density of the data for a particular class is significantly higher than its density in the overall data.

| Data set | Dim. $(d)$ | Size $(n)$ | Classes | Class 1 | Class 2 | Class 3 |
|----------|-----------|-----------|---------|---------|---------|---------|
| *Ionosphere* | 2 | 351 | 2 | 225 | 126 | – |
| *Haberman* | 3 | 306 | 2 | 225 | 81 | – |
| *Iris* | 4 | 150 | 3 | 50 | 50 | 50 |
| *Transfusion* | 4 | 748 | 2 | 570 | 178 | – |

Table I.    Datasets employed in experiments.

It must be noticed that none of these methods investigates the extension of the nearest neighbor decision rule to the handling of uncertain data. Moreover, in the experimental section comparison between UNN and density based methods for classification will be investigated.

## 5. EXPERIMENTAL RESULTS

This section presents results obtained by experimenting the UNN rule.

Experiments are organized as follows. Section 5.2 studies the effect of disregarding data uncertainty on classification accuracy. Section 5.3 investigates the behavior of UNN on test objects whose label is independent of the theoretical prediction and its sensitivity to noise. Section 5.4 reports execution time by using both certain and uncertain test objects. Section 5.5 compares the approach here proposed with density based classification methods for uncertain data. Section 5.6 describes a real-life scenario in which data are naturally modelled as multi-dimensional pdfs.

First of all, the following section describes the characteristics of some of the datasets employed in the experimental activity.

### 5.1    Datasets description

Table I reports datasets employed in the experiments and their characteristics. All the datasets are from the UCI ML Repository [Asuncion and Newman 2007]. As for the *Ionosphere* dataset, it has been projected on the two principal components.

For each dataset above listed, a family of uncertain training sets has been obtained. Each training set of the family is characterized by a parameter $s$ (for *spread*) used to determine the degree of uncertainty associated with dataset objects. In particular, for each certain object $x_i = (x_{i,1}, \ldots, x_{i,d})$ in the original dataset, an uncertain object $x_i'$ has been associated with, having pdf $f^i(v_1, \ldots, v_d) = f_1^i(v_1) \cdot \ldots \cdot f_d^i(v_d)$. Each one dimensional pdf $f_j^i$ is randomly set to a normal or to a uniform distribution, with mean $x_{i,j}$ and support $[a, b]$ depending on the parameter $s$. In particular, let $r$ be a randomly generated number in the interval $[0.01 \cdot s \cdot \sigma_j, s \cdot \sigma_j]$, where $\sigma_j$ denotes the standard deviation of the dataset along the $j$th coordinate, then $a = x_{i,j} - 4 \cdot r$ and $b = x_{i,j} + 4 \cdot r$.

In the experiments, the parameter $N$, determining the resolution of integrals,

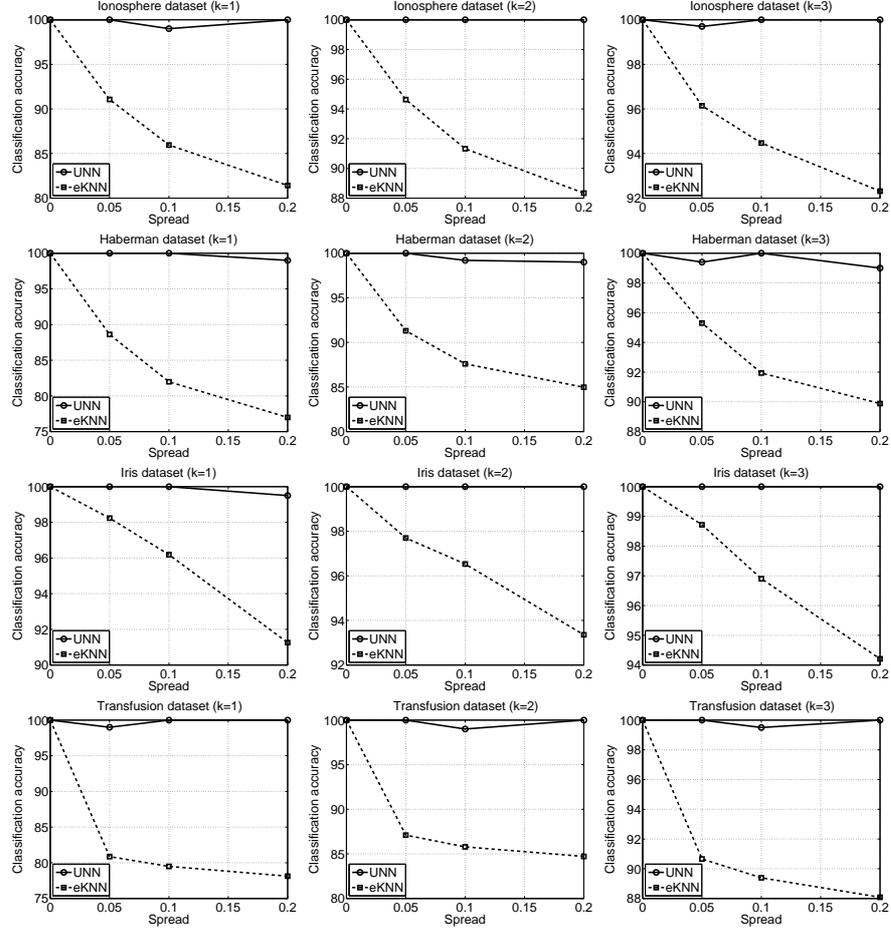

Fig. 4.   Accuracy using random certain queries.

has been set to $N = 100 \cdot 2^d$, while the histogram resolution $h$ has been set to 100. Furthermore, experimental results are averaged on ten runs.

## 5.2   The effect of disregarding uncertainty

The goal of this experiment is to show that whenever uncertain data are available, taking into account uncertainty leads to superior classification results.

With this aim, two algorithms have been implemented to be compared with UNN, namely the *Random* and the eKNN (for expected $k$-nearest neighbor) algorithms. The *Random* algorithm approximates the expression reported in Equation (1) by randomly generating $M$ outcomes $I_T$ of the uncertain training set $T$, and, hence, determines the most probable class of the test object (see Equation (2)).  The eKNN algorithm randomly generates $M$ outcomes $I_T$ of the uncertain training set $T$, classifies test objects by applying the $k'$ nearest neighbor rule with training set $I_T$ ($k'$ is set to $2k-1$, according to Proposition 2.6) and, finally, reports the average

| | $s = 0.05$ | | | $s = 0.10$ | | | $s = 0.20$ | | |
|---|---|---|---|---|---|---|---|---|---|
| | $k = 1$ | $k = 2$ | $k = 3$ | $k = 1$ | $k = 2$ | $k = 3$ | $k = 1$ | $k = 2$ | $k = 3$ |
| *Spiral* | 63.3 | 63.2 | 66.2 | 59.9 | 62.4 | 64.8 | 56.9 | 59.9 | 62.1 |
| *Ionosphere* | 78.6 | 73.7 | 73.1 | 68.7 | 67.9 | 68.6 | 62.9 | 66.5 | 71.2 |
| *Haberman* | 83.5 | 78.0 | 76.8 | 73.0 | 70.1 | 70.8 | 66.4 | 68.8 | 73.6 |
| *Iris* | 91.9 | 87.2 | 88.1 | 83.6 | 81.8 | 83.8 | 75.3 | 78.4 | 82.2 |
| *Transfusion* | 75.1 | 83.0 | 87.6 | 74.2 | 82.2 | 87.0 | 74.0 | 82.2 | 87.0 |

Table II.   Accuracy of eKNN on border test objects.

classification accuracy over all the outcomes. [2]

For each uncertain training set, one thousand certain test objects have been randomly generated. The generic test object $q$ is obtained as $q = (x_i + x_j)/2$, where $x_i$ and $x_j$ are two randomly selected certain dataset objects.

The label reported by the *Random* algorithm, has been employed as its true label. Hence, the accuracy of the UNN classification algorithm has been compared with the accuracy of the eKNN algorithm on the test set. Since this experiment computes the accuracy of the eKNN algorithm with respect to the theoretical prediction, it determines how the certain nearest neighbor rule is expected to perform over a generic outcome of the uncertain dataset. In other words, the experiment *measures the accuracy of the classification strategy based on disregarding data uncertainty*, which is the approach of encoding each (uncertain) object by means of one single measurement and then employing the certain nearest neighbor rule to perform classification. This accuracy is moreover compared with that of the uncertain nearest neighbor rule which, conversely, takes into account the underlying uncertain data distribution.

Figure 4 shows the accuracy of UNN and eKNN methods for various values of spread ($s \in [0, 0.20]$) and $k$ ($k \in \{1, 2, 3\}$). It is clear that the accuracy of UNN is very high for all spreads, in that it is almost always close to 100%. There are some discrepancies with the theoretical prediction, whose number slightly increases with the uncertainty in the data, due to the fact that approximate computations are employed by both the *Random* and the UNN algorithm.

As far as the eKNN algorithm is concerned, it is clear from the results that its prediction may be very inaccurate. In particular, the greater the level of uncertainty in the data, the smaller its accuracy. Recall that for certain datasets (that is, for spread $s = 0$), the two classification rules coincide. In the experiments, the difference in accuracy of the eKNN with respect to the UNN can reach the 20%, in correspondence of the largest value of spread considered.

As for the effect of the parameter $k$, it appears that the accuracy of eKNN gets better with larger values of $k$, though it remains unsatisfactory in all cases. This behavior can be justified by considering the rule used to generate test objects. These objects represent the mean of two randomly selected points, hence a large fraction of them lie outside the decision boundary. For test objects sorrounded by objects of the same class, the majority vote tends to approximate the most probable class, and this is particularly true for small spreads, since the region where these test objects are located tends to present non-null probability for only one of the two classes.

---

[2]The parameter $M$ has been set equal to the number of points used to compute integrals, that is either $N$, for certain queries, or $N^2$, for uncertain ones.

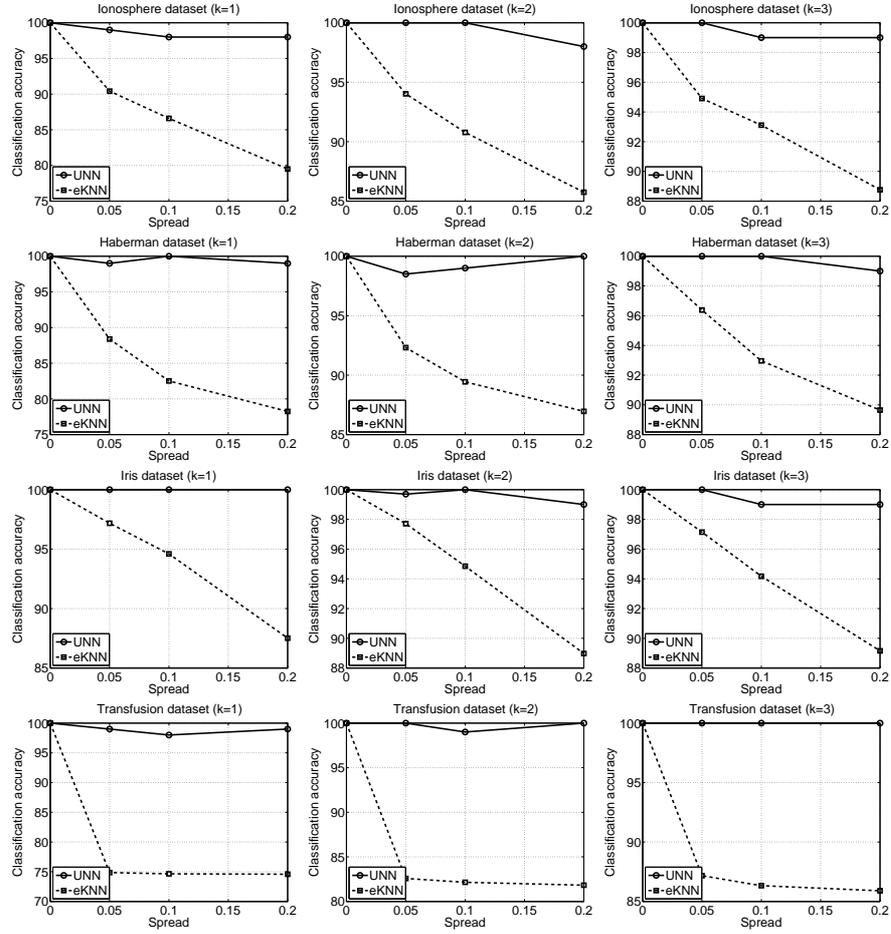

Fig. 5. Accuracy using random uncertain queries.

In order to study the behavior on critical test objects, that are objects located along the decision boundary, the above experiment was repeated on a further set of one thousand test objects, called *border* test objects, determined as explained next. The generic border test object $q$ is obtained as $q = (x_i + x_j)/2$, where $x_i$ and $x_j$ are two randomly selected certain dataset objects and $q$ satisfies the condition that the mean distances $d_c^q = \frac{1}{k} \sum_i nn_i(q, T_c)$ and $d_{c'}^q = \frac{1}{k} \sum_i nn_i(q, T_{c'})$ are similar (namely their difference is within the ten percent), that is $|d_c^q - d_{c'}^q| / \max\{d_c^q, d_{c'}^q\} \le 0.1$.

On these objects, the behavior of UNN is similar to that exhibited on the random test objects. Table II reports the accuracy of eKNN on the border test objects for the various values of spread $s$ and nearest neighbors $k$. It is clear that the accuracy of eKNN further deteriorates: the accuracy may decrease of an additional 20% percent with respect to the previous experiment. Moreover, the advantage of increasing the value of $k$ becomes less evident. In some cases the accuracy does not vary with $k$ or may even get worse for larger values.

Figure 5 shows the accuracy of UNN and eKNN for random uncertain test objects. The uncertain test objects have been obtained by centering multi-dimensional pdfs, generated according to the policy used for the training set objects, on the certain test objects employed in the experiment of Figure 4. The trend of these curves is similar to that associated with curves obtained for certain test objects. Particularly, in many cases the accuracy of eKNN worsens by some percentage points with respect to the certain test objects. This can be explained since in this experiment the data uncertainty has increased.

Concluding, the experimental results presented in this section confirm that classification results benefit from taking into account data uncertainty.

### 5.3 Experiments on real labels and robustness to noise

In this experiment the accuracy of the UNN, the eKNN, and the certain $k$ nearest neighbor algorithm (referred to as KNN in the following) has been compared by taking into account the original dataset labels of the test objects.

The range of values for the spread $s$ and for the number of nearest neighbors $k$ considered are $s \in [0, 0.2]$ and $k \in \{1, 2, 3\}$, respectively, which are the same employed in the experiment described in the previous section. UNN and eKNN have been executed on the uncertain version of the dataset, while KNN has been executed on the certain dataset. Accuracy has been measured by means of ten fold cross validation.

Note that, while the certain dataset can be assimilated to a generic outcome of an hypothetical true uncertain dataset which is unknown, the uncertain dataset here employed has been syntetically generated by using arbitrary distributions centered on the certain dataset objects (as already explained at the beginning of the experimental result section) and it is not intended to represent the (unknown) true uncertain dataset.

Thus, it is important to point out that the purpose of this experiment is neither to demonstrate that that UNN peforms better than KNN (as a matter of fact the two methods are designed for two very different application scenarios; recall that UNN is executed on uncertain data, while KNN can be executed only on certain data) nor to show that better classification results can be achieved by injecting uncertainty in the data. Rather, the goal of the experiment is to study the behavior of UNN on test objects whose label is independent of the theoretical prediction and, particularly, to appreciate the sensitivity of UNN to noise. With this aim, the accuracy of KNN will be employed as a baseline to assess the accuracy of UNN, since the output of KNN represents the classification achieved on the considered datasets by the nearest neighbor classification rule when uncertainty disappears.

Figure 6 shows the result of the experiment. Curves report the accuracy of UNN (solid lines), eKNN (dashed lines), and KNN (dotted lines). On the *Ionosphere*, *Haberman*, and *Transfusion* datasets the accuracy of UNN is above than that of KNN. Moreover, on the two latter datasets, the accuracy is slightly increasing with the data uncertainty (spread). The difference in accuracy can be justified by noticing that UNN mitigates the effect of noisy points since it takes simultaneously into account the whole class probability, according to the theoretical analysis depicted in Section 2. As for the *Iris* dataset, the accuracy of UNN is practically the same as that of KNN. This can be justified since this dataset contains a little amount of noise and it is composed of well-separated classes.

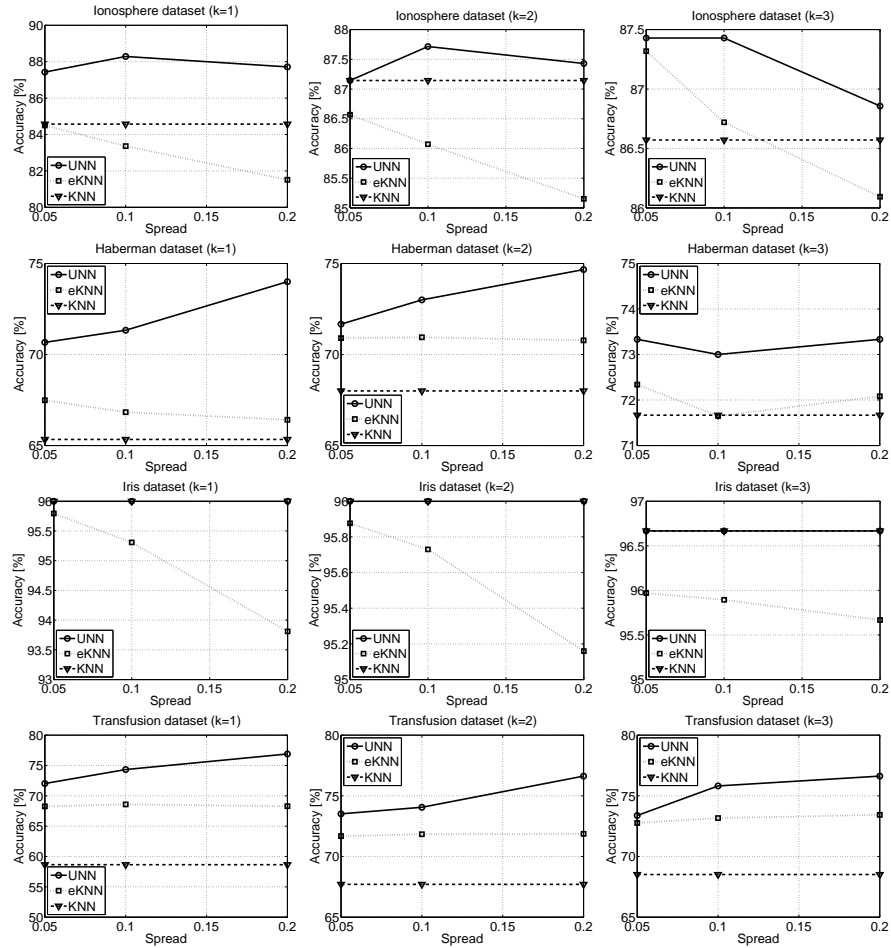

Fig. 6.    Ten-fold cross validation results.

As far as the comparison of UNN and eKNN is concerned, the former method performs always better than the latter, thus confirming the result of the analysis conducted in the previous section. As for the effect of the parameter $k$ on the accuracy, as already discussed the accuracy of eKNN improves for larger values of $k$. However, it is well-known that it is difficult to select a nearly optimum value of $k$ to approach the lowest possible probability of error. In particular, as $k$ increases beyond a certain value, which depends on the nature of the dataset, the probability of error may begin to increase. The plots show that UNN achieves very good results by using the smallest possible value for $k$, that is $k = 1$, and that in different cases the maximum accuracy is achieved for values of $k$ smaller than the greatest value here considered (e.g., see *Ionosphere* for $k = 1$ and $s = 0.1$, or *Haberman* for $k = 2$

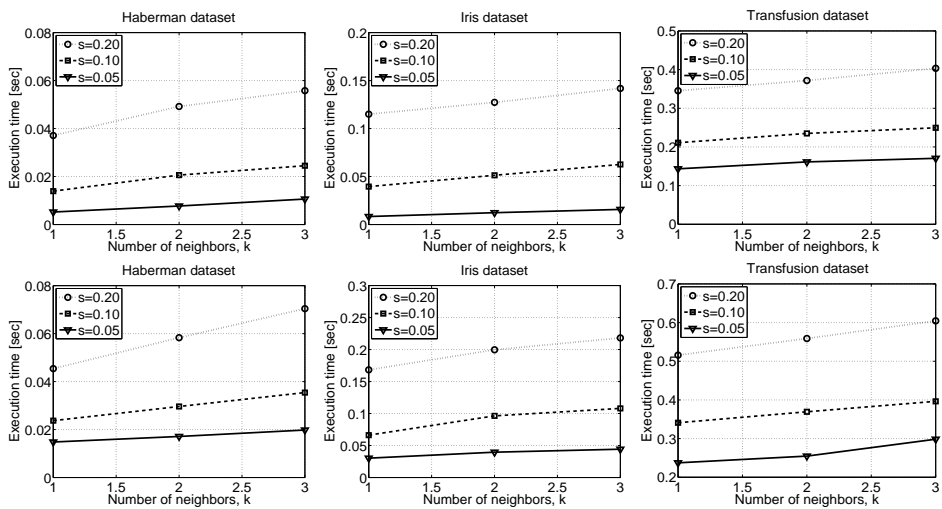

Fig. 7.    UNN execution time.

and $s = 0.2$)

Thus, these experimental results confirm the discussion of Section 2, where it is pointed out that the concept of nearest neighbor class is more powerful than that of nearest neighbor in presence of uncertainty.

### 5.4    Execution time

Figure 7 reports the time employed by UNN to classify one single test object.[3]

Plots in the first (second, resp.)  row of Figure 7 show the execution time on the *Haberman*, *Iris*, and *Transfusion* datasets when certain (uncertain, resp.) test objects are employed. Clearly, the execution time increases both with $k$ and with the data uncertainty: The larger the spread, the greater the execution time; moreover, classifying uncertain test objects requires more time than classifying certain ones. Indeed, the larger the uncertainty ($k$, resp.), the larger the radius $R_{max}^q$ and, consequently, the number of integrals to be computed.

The following table reports the *relative execution time* of UNN, that is the ratio between the execution time of the UNN algorithm (which computes the integral in Equation (15)) and the time needed to compute the integral in Equation (7) when all the training set objects are taken into account. Thus, the table shows the time savings obtained by exploiting techniques reported in Section 3.

| Test set | Dataset | $s = 0.05$ | | | $s = 0.10$ | | | $s = 0.20$ | | |
|---|---|---|---|---|---|---|---|---|---|---|
| | | $k=1$ | $k=2$ | $k=3$ | $k=1$ | $k=2$ | $k=3$ | $k=1$ | $k=2$ | $k=3$ |
| *Certain* | *Haberman* | 0.01 | 0.02 | 0.02 | 0.03 | 0.05 | 0.05 | 0.09 | 0.11 | 0.13 |
| | *Iris* | 0.01 | 0.01 | 0.01 | 0.04 | 0.05 | 0.06 | 0.11 | 0.12 | 0.13 |
| | *Transfusion* | 0.05 | 0.06 | 0.06 | 0.08 | 0.09 | 0.09 | 0.13 | 0.14 | 0.15 |
| *Uncertain* | *Haberman* | 0.08 | 0.09 | 0.10 | 0.13 | 0.16 | 0.19 | 0.25 | 0.31 | 0.37 |
| | *Iris* | 0.03 | 0.04 | 0.05 | 0.07 | 0.11 | 0.12 | 0.19 | 0.22 | 0.24 |
| | *Transfusion* | 0.11 | 0.12 | 0.13 | 0.15 | 0.17 | 0.18 | 0.23 | 0.25 | 0.27 |

---

[3]Experiments were executed on a CPU Core 2 Duo 2.40GHz with 4GB of main memory under the Linux operating system.

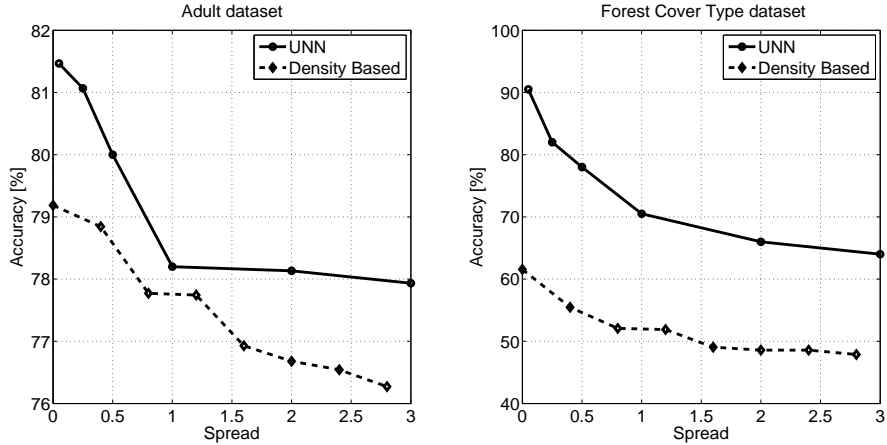

Fig. 8.   Comparison with the Density Based classification algorithm.

The relative execution times reported in the table show that properties exploited by UNN to accelerate computation guarantee time savings in all cases. For certain test objects, in most cases the relative execution time is approximatively below 0.10, and in some cases it is even close to 0.01, e.g., see the *Haberman* and *Iris* datasets. Also for uncertain test objects, in many cases it is approximatively below 0.15, though in different cases it is much smaller.

For spread $s = 0.2$ a considerable fraction of the dataset objects are within distance $R^q_{max}$ from the test object $q$ and, hence, the relative execution time increases. This effect is more evident when uncertain test objects are taken into account. However, it can be noted that the spread $s = 0.2$ is very large, in fact in this case the supports of the training set objects are rather wide and tend to partially overlap.

### 5.5   Comparison with density based methods

This section describes comparison between UNN and the *Density Based* classification algorithm proposed in [Aggarwal 2007], where a general framework for dealing with uncertain data is presented.

The same experimental setting described in [Aggarwal 2007] is considered. Following the methodology therein proposed, an uncertain dataset is generated starting from a certain one, as described next. First of all, just the numerical attributes are taken into account, let $d$ be their number. Then, for each object $x_i = (x_{i,1}, \ldots, x_{i,d})$ in the original certain dataset, an uncertain object $x'_i$ with pdf $f^i(v_1, \ldots, v_d) = f^i_1(v_1) \cdot \ldots \cdot f^i_d(v_d)$ is generated. Each one dimensional pdf $f^i_j$ is a normal distribution with mean $x_{i,j}$ and standard deviation equal to $[0, 2 \cdot s \cdot \sigma_j]$, where $\sigma_j$ is the standard deviation of the dataset objects along the $j$-th attribute. Thus, the value of the spread $s$ determines the uncertainty level of the dataset and has been varied in the range $[0, 3]$.

Two datasets coming from the UCI ML Repository [Asuncion and Newman 2007], that are *Adult* and *Forest Cover Type*, are employed. The former contains data extracted from the census bureau database. It consists of 32,561 objects and six numerical attributes. The latter dataset contains data about forest cover type of four areas located in the northern Colorado. It consists of 581,012 objects and ten numerical attributes.

Figure 8 reports experimental results. In all cases UNN exhibited a better classification accuracy than the Density Based algorithm. The accuracy of both methods degrades with the spread $s$. It can be noticed that the difference between the case $s = 0$ and the case $s = 3$ is substantial for both methods, and this can be justified by noticing that in the latter case the level of uncertainty is very high, with a lot of dataset objects having overlapping domain. However, UNN shows itself to be sensibly more accurate for all levels of uncertainty, thus confirming the effectiveness of the concept of nearest neighbor class.

### 5.6    A real-life example application scenario

This section describes a real-life prediction scenario in which data can be naturally modelled by means of multi-dimensional continuous pdfs, that is the most general form of uncertain objects managed by the technique here introduced, and illustrates the meaningfulness of uncertain nearest neighbor classification within the described task.

The scenario concerns Mobile Ad hoc NETworks (or MANETs). A MANET [Bai and Helmy 2006] is a collection of wireless mobile nodes forming a self-configuring network without using any existing infrastructure. Potential applications of MANETs are mobile classrooms, battlefield communication, disaster relief, and others. The *mobility model* of a MANET is designed to describe the movement pattern of mobile users, and how their location, velocity and acceleration change over time. One frequently used mobility model in MANET simulations is the *Random Waypoint* model [Broch et al. 1998], in which nodes move independently to a randomly chosen destination with a randomly selected velocity within a certain simulation area. For such a model, the spatial node distribution is such that the node density is maximum at the center of the simulation area, whereas the density is almost zero around the boundary of the area, hence the distribution is non-uniform. Moreover, no matter how fast the nodes move, the spatial node distribution at a certain position is only determined by its location [Bettstetter et al. 2004]. For a squared area of size $a$ by $a$, centered in $(x_0, y_0)$, the pdf of the random waypoint model is provided by the following analytical expression:

$$f_{\mathrm{rw}}(x, y) \approx \frac{36}{a^6} \cdot \left( (x - x_0)^2 - \frac{a^2}{4} \right) \cdot \left( (y - y_0)^2 - \frac{a^2}{4} \right),$$

for $x \in \left[ x_0 - \frac{a}{2}, x_0 + \frac{a}{2} \right]$ and $y \in \left[ y_0 - \frac{a}{2}, y_0 + \frac{a}{2} \right]$, and $f_{\mathrm{rw}}(x, y) = 0$ outside. Figure 9 shows the function $f_{\mathrm{rw}}$.

In such networks the nodes may dynamically enter the network as well as leave it. The nodes of a MANET are typically distinguished by their limited power, processing, and memory resources, as well as high degree of mobility. Since nodes are not able to be re-charged in an expected time period, energy conservation is crucial to maintaining the life-time of nodes. One of the goals of protocols is to minimize energy consumption through techinques for routing, for data dissemination, and for varying transmission power (and, consequently, transmission range). Multiple hops are usually needed for a node to exchange information with any other node in the network, and nodes take adavantage of their neighbors in order to communicate with the rest of the network nodes.

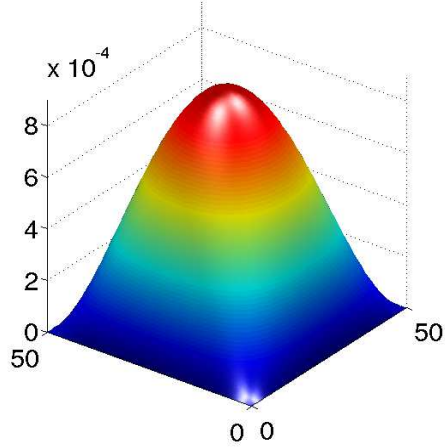

Fig. 9.   The pdf $f_{\mathrm{rw}}$ of the Random waypoint model for MANETs.

As a matter of fact, the needed transmission power is inversely proportional to the squared distance separating the transmitter to the receiver [Wesolowski 2002]. For an isotropic antenna the radiation $P_r$ at a distance $R$ is

$$P_r(R) = \frac{P_t}{R^\alpha},$$

where $P_t$ is the transmitted signal strength and $\alpha$ is the path loss factor, which depends on the given propagation environment and whose value is typically between 2 (in free space) and 6. Since a node can correctly receive packets if the signal strength $P_r$ of the packet at that node is above a certain threshold, and since mobile devices exploit variable-range transmission as a powersave strategy, the minimum power to be supplied by a node $v$ connected to the network $W$ is:

$$pow_W(v) \propto \int_0^{+\infty} R^\alpha \cdot Pr(\mathrm{d}(v, nn_W(v)) = R) \, \mathrm{d}R. \qquad (21)$$

Thus, pdfs $f_{\mathrm{rw}}$ naturally model uncertain objects representing mobile devices, also called *nodes* in the following. In the experiment described in this section, an uncertain training set of nodes partitioned in two classes, representing two different MANET networks, is considered. The simulation area is the unit square centered in the origin. The red network has ten nodes randomly positioned in the whole simulation area and allowed to move in squares of size 0.2 by 0.2 (these centers are identified in Figure 10(a) by means of plus-marks), while the blue network has five nodes randomly positioned in the lower-right corner of the simulation area and allowed to move in squares of size 0.05 by 0.05 (these centers are identified in Figure 10(a) by means of x-marks). Certain objects are the points of the plane. A set of 2,500 randomly generated points within the simulation square has been employed as test set. These points have been labelled as either red or blue depending on the minimum achievable power consumption of the node, namely the label of the certain test object $q$ is determined as

$$\arg \min_{W \in \{red, blue\}} pow_W(q).$$

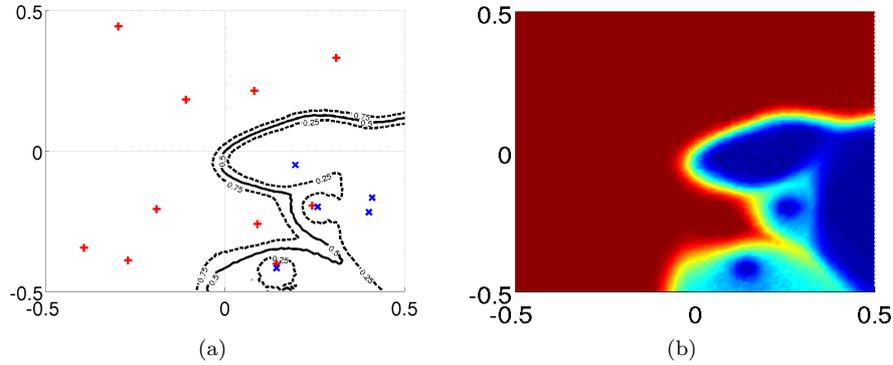

Fig. 10.   Experimental results on the MANET training set.

Thus, the classification task considered consists in the prediciton of the less demanding network in terms of energy to be expended.

The UNN applied to the above described dataset returns the nearest neighbor class of a test object, that is to say the network (class) that minimizes the expected distance from the position of a node determined to join a neighborhood MANET (the test object) and the uncertain position of one of its nodes (members).

Figure 10 reports the classification of the points of the plane for $k = 1$. In Figure 10(b) points are colored according to the probability to belong to one of the two classes. In Figure 10(a) the solid black curve represents the decision boundary, that is the points for which the nearest neighbor class probability equals 0.5. The two dashed curves correspond to the points having red class probability 0.75 and 0.25. The form of the decision boundary is informative, since it differs from the common facets of the adjacent Voronoi cells associated with objects belonging to opposite classes, that is the decision boundary of the certain nearest neighbor rule. In particular, it can be observed that the centers of two red nodes are within the support of the blue class (and that the probability of the red class is below 0.25 in correspondence of both these two centers). This can be justified by noticing that these two centers are close to the centers of two blue nodes, and that the mobility of blue nodes is smaller than that of red ones.

The accuracy of UNN on the test set has been measured and compared with that of eKNN. The accuracy of UNN was 0.986, while that of eKNN was 0.938. The good performance of nearest neighbor based classification methods is due to the fact that power consumption is related to the Euclidean distance between devices. Note that, while the uncertain nearest neighbor rule reports the class which most probably provides the nearest neighbor, the power (Equation (21)) depends also on the distribution of the distance separating the transmitter to its nearest neighbor, and this explains why there are misclassifications. Also in this experiment, UNN performs better than eKNN, and this can be explained since the former rule bases its decision on the concept of nearest neighbor class, thus confirming the superiority of the uncertain nearest neighbor rule even with respect to classical classification techniques in presence of uncertainty.

## 6. CONCLUSIONS

In this work the uncertain nearest neighbor rule, representing the generalization of the certain nearest neighbor rule to the uncertain scenario, has been introduced. It has been provided evidence that the uncertain nearest neighbor rule correctly models the right semantics of the nearest neighbor decision rule when applied to the uncertain scenario. Moreover, an algorithm to perform uncertain nearest neighbor classification of a generic (un)certain test object has been presented, together with some properties precisely designed to significantly reduce the temporal cost associated with nearest neighbor class probability computation. The theoretical analysis and the experimental campaign here presented have shown that the proposed algorithm is efficient and effective in classifying uncertain data.